%% file: main.tex
\documentclass[letterpaper, 10 pt, conference]{ieeeconf}
\IEEEoverridecommandlockouts
\usepackage{comment}
\usepackage{amsmath}

\usepackage{graphicx}
\usepackage{rotating}
\usepackage{float}
\usepackage{mathrsfs}
\usepackage{amssymb}
\usepackage{autobreak}
\usepackage{mathtools}
\usepackage{wrapfig}
\usepackage{bbm}
\usepackage{algorithm}
\usepackage{graphicx, subcaption}
\usepackage{makecell}
\usepackage{booktabs}
\usepackage[svgnames]{xcolor}
\usepackage[normalem]{ulem} 
\usepackage{gensymb}
\usepackage[]{algpseudocode}

\usepackage{lipsum}
\usepackage{outlines}
\usepackage{blindtext}
\usepackage{multicol}
\usepackage{tikz}
\usetikzlibrary{shapes,backgrounds}
\usepackage{array}
\usepackage{wrapfig}
\usepackage{thmtools} 
\usepackage{thm-restate}
 
\usepackage{epsfig} 
\usepackage{amsmath} 
\usepackage{algorithm}
\usepackage[]{algpseudocode}
\usepackage{tablefootnote}

\def\secref#1{Sec.~\ref{#1}}
\def\figref#1{Fig.~\ref{#1}}
\def\tabref#1{Tab.~\ref{#1}}
\def\eqref#1{Eq.~(\ref{#1})}
\def\algref#1{Alg.~\ref{#1}}
\def\ie{\emph{i.e.}}
\def\eg{\emph{e.g.}}
\def\etal{\emph{et al. }}

\usepackage{glossaries-extra}
\setabbreviationstyle[acronym]{long-short}
\glssetcategoryattribute{acronym}{nohyperfirst}{true}

\makeglossaries

\newacronym{slam}{SLAM}{Simultaneous Localization and Mapping}
\newacronym{ba}{BA}{Bundle Adjustment}
\newacronym{sfm}{SfM}{Structure from Motion}
\newacronym{pgo}{PGO}{Pose-Graph Optimization}
\newacronym{vpr}{VPR}{Visual Place Recognition}
\newacronym{sgd}{SGD}{Stochastic Gradient Descent}
\newacronym{ils}{ILS}{Iterative Least-Squares}
\newacronym{gn}{GN}{Gauss-Newton}
\newacronym{lm}{LM}{Levenberg-Marquardt}
\newacronym{sdp}{SDP}{Semi-Definite Programming}
\newacronym{vo}{VO}{Visual Odometry}
\newacronym{vio}{VIO}{Visual-Inertial Odometry}
\newacronym{imu}{IMU}{Inertial Measurement Unit}
\newacronym{pnp}{PnP}{Perspective-n-Point}
\newacronym{dof}{DoF}{Degrees of Freedom}
\newacronym{ar}{AR}{Augmented Reality}
\newacronym{sota}{SOTA}{state-of-the-art}
\newacronym{rpr}{RPR}{Relative Pose Regression}

\def\sota{\gls{sota} }
\def\slam{\gls{slam} }
\def\sfm{\gls{sfm} }
\def\dof{\gls{dof} }
\def\rpr{\gls{rpr} }

\input{macros.tex}
\title{Augmented Reality without Borders: Achieving Precise Localization Without Maps}
\author{Albert Gassol Puigjaner, Irvin Aloise and Patrik Schmuck
\thanks{All authors are with Magic Leap, Switzerland {\tt albertgassol1@gmail.com}, {\tt\{ialoise,pschmuck\}@magicleap.com}.}}%

\makeatletter
\let\NAT@parse\undefined
\makeatother
\definecolor{lightblue}{rgb}{0.12,0.49,0.85}
\usepackage[colorlinks=true,linkcolor=black,citecolor=blue,urlcolor=blue]{hyperref} 
\usepackage[capitalise]{cleveref}
\crefname{assumption}{Assumption}{Assumptions}

\usepackage{tikz}

\begin{document}

\maketitle
\thispagestyle{empty}
\pagestyle{empty}


\begin{abstract}
Visual localization is crucial for Computer Vision and \gls{ar} applications, where determining the camera or device's position and orientation is essential to accurately interact with the physical environment. Traditional methods rely on detailed 3D maps constructed using \sfm or \gls{slam}, which is computationally expensive and impractical for dynamic or large-scale environments. We introduce \method, a novel localization framework for \gls{ar} applications that uses known relative transformations within image sequences to perform intra-sequence triangulation, generating 3D-2D correspondences for pose estimation and refinement. \method eliminates the need for pre-built \sfm maps, providing accurate and efficient localization suitable for dynamic outdoor environments. Evaluation with benchmark datasets and real-world experiments demonstrates \method's state-of-the-art performance and robustness. By integrating \method into an \gls{ar} device, we highlight its capability to achieve precise localization in real-world outdoor scenarios, showcasing its practical effectiveness and potential to enhance visual localization in \gls{ar} applications.
\end{abstract}



\setlength{\textfloatsep}{5pt}
\input{1-intro}

\input{2-related_work}

\input{3-background}

\input{4-method}
\input{5-experimental_results}
\input{6-conclusions}

\bibliographystyle{IEEEtran}
\bibliography{main}

\end{document}

%% file: macros.tex
\input{notation}

\newcommand{\mypar}[1]{\noindent\textbf{#1}.}
\newcommand{\method}{\textsc{\small{MARLoc}}\xspace}

\newcommand{\sethree}{\mathrm{SE}(3)}
\newcommand{\sothree}{\mathrm{SO}(3)}
\newcommand{\pose}[2]{{}_{{#1}}\bT_{{#2}}}
\newcommand{\rot}[2]{{}_{{#1}}\bR_{{#2}}}
\newcommand{\quaternion}[2]{{}_{{#1}}\bq_{{#2}}}
\newcommand{\translation}[2]{{}_{{#1}}\bt_{{#2}}}

\newcommand{\queryPose}[1]{{}_{q}\bT_{{#1}}^{Q}}
\newcommand{\querySequence}{\cI_q}
\newcommand{\queryPoses}{\cT_q}
\newcommand{\queryImage}[1]{I_{{#1}}^q}

\newcommand{\neighborSequence}{\cN_q}

\newcommand{\refSequence}{\cI_r}
\newcommand{\refPose}[1]{{}_{r}\bT_{{#1}}^{R}}
\newcommand{\refPoses}{\cT_r}
\newcommand{\refImage}[1]{I_{{#1}}^r}

\newcommand{\localizedPose}[2]{{}_{r}\bT_{{#1}}^{Q{#2}}}

\newcommand{\keypoint}[3]{\bp^{{#1}}_{{#2},{#3}}}
\newcommand{\liftedKeypoint}[3]{\bP^{{#1}}_{{#2},{#3}}}
\newcommand{\correspondences}{\cM}

\newcommand\boldblue[1]
{\textcolor{RoyalBlue}{\textbf{#1}}}
\newcommand\boldgreen[1]{\textcolor{Green}{\textbf{#1}}}

%% file: notation.tex
\usepackage{amsopn}

\newcommand{\bt}{\mathbf{t}}

\newcommand{\bR}{\mathbf{R}}

\newcommand{\bT}{\mathbf{T}}

\newcommand{\cN}{\mathcal{N}}
\newcommand{\cM}{\mathcal{M}}

\newcommand{\cI}{\mathcal{I}}
\newcommand{\cT}{\mathcal{T}}

\newcommand{\bbR}{\mathbb{R}}

\newcommand{\be}{\mathbf{e}}

\newcommand{\bq}{\mathbf{q}}
\newcommand{\bP}{\mathbf{P}}

\newcommand{\bp}{\mathbf{p}}

\def\g2o{$g^2o$}
\def\t2v{\mathrm{t2v}}
\def\v2t{\mathrm{v2t}}
\def\ev2t{\mathrm{ev2t}}

%% file: 1-intro.tex

\section{Introduction}\label{sec:intro}

\glsreset{ar}
\gls{ar} is a relatively new technology, and the research community is investigating ways to leverage this technology in several fields, ranging from medical applications over navigation to video games.
One of the main characteristics that makes \gls{ar} so appealing is its ability to seamlessly blend digital content with the real world. This allows users to interact simultaneously with the physical environment and its digitally enhanced counterpart.
Consequently, the ability to accurately distribute \emph{world-locked} virtual content in physical environments is extremely valuable for successfully creating \gls{ar} applications.

In this context, the well-established topic of Visual Localization is a fundamental building block for creating such world-locked immersive and interactive \gls{ar} experiences, since it allows to exactly determine the position of the \gls{ar} device within the environment. 
With Visual Localization, we refer to the Computer Vision problem of determining the position and orientation of an image with respect to another image. 
Traditional methods solve this problem by relying on a digital reconstruction of the geometry of the environment, 
typically resorting to techniques such as \sfm~\cite{schoenberger2016sfm} or \slam~\cite{thrun2005probabilistic,grisetti2010graphslam,murTRO2015} 
to construct pointclouds used at localization time~\cite{schoenberger2016mvs,Sattler2012ImprovingIL,sarlin2019hloc}.
This family of approaches has been proven to be very effective, however, they also come with several shortcomings.
As the main drawback, creating a digital representation of the geometry of the environment (commonly referred to as a \emph{map}) is a computationally demanding and time-consuming operation.
To have some grounding, an SfM-based reconstruction of an office-like environment can take hours to complete on a powerful workstation~\cite{brachmann2023accelerated}, 
which can constitute a problem for interactive \gls{ar} applications where compute resources are rather limited and impose a long wait for the user to interact with the \gls{ar} experience.
Furthermore, the images employed in the processing are required to have a good overlap for the reconstruction to be successful.
Faster ways of creating maps from a sequence of images exist in the literature; however, approaches like \slam usually require additional sensor inputs - \eg~\gls{imu}~\cite{delmerico2018viobenchmark,Bloesch2015rovio,Forster2015OnManifoldPF} - to succeed
and often result in very sparse maps, which are not optimal for visual localization algorithms.
Finally, oftentimes it is not practical to manually capture images for the reference environment, making it complex to perform any sort of prior reconstruction, such as when operating in large outdoor environments or deploying AR experiences in distant places.
Given this, the research community is investigating ways of generating accurate visual localization estimates in \emph{map-free} setups~\cite{arnold2022mapfree}, only using a sparse set of images as a digital representation of the reference environment.
This setup allows us to avoid performing expensive mapping sessions to prepare the localization framework and streamlines the overall integration in real-time applications such as \gls{ar}.

\begin{figure}[t]
    \centering
    \includegraphics[width=1\columnwidth]{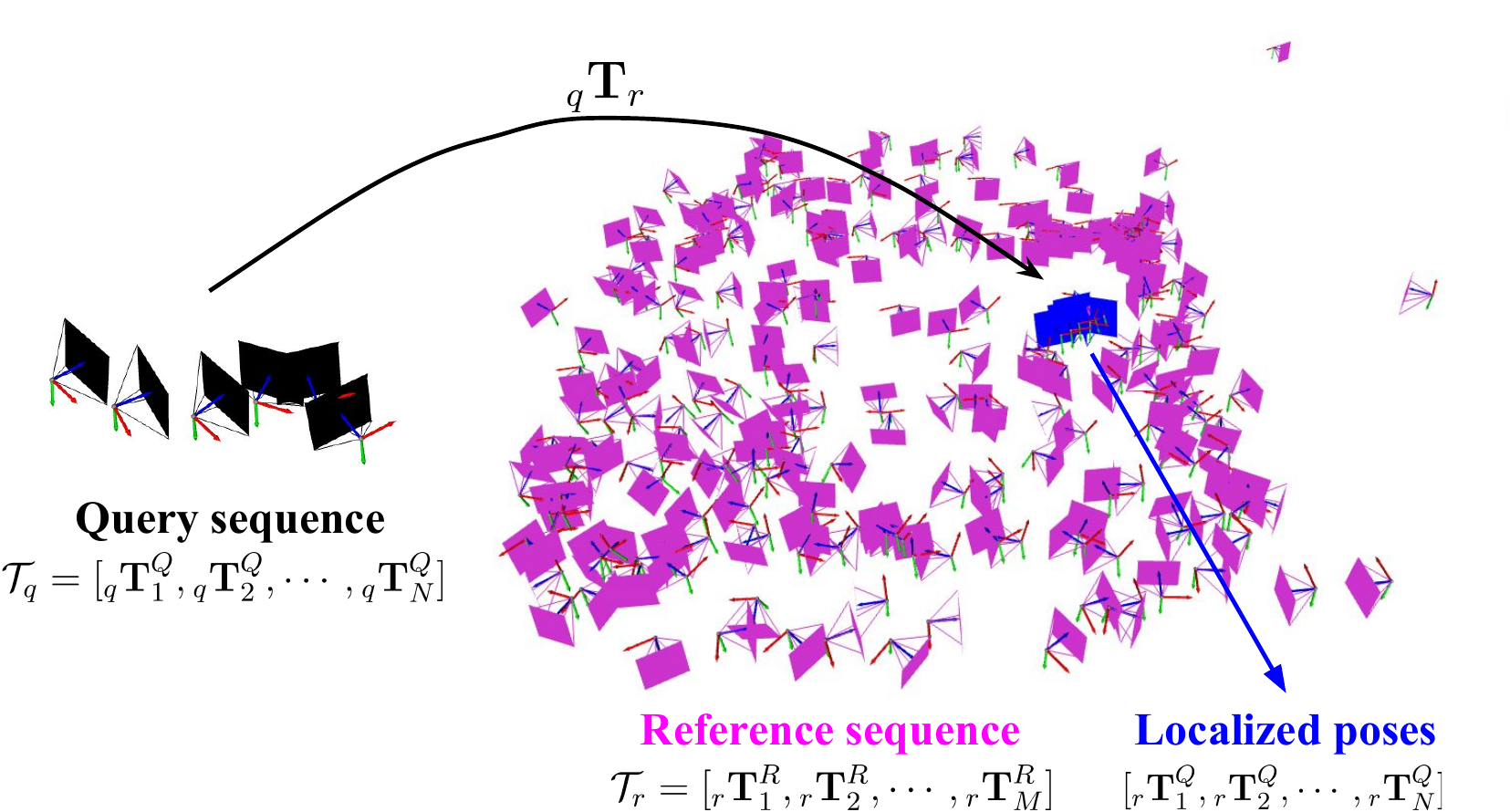}
    \caption{Given a sequence of \textcolor{black}{\textbf{locally posed query frames}} $\queryPoses$ and a set of \textcolor{Fuchsia}{\textbf{posed reference frames}} $\refPoses$, our objective is to find $\{\localizedPose{1}{}, \localizedPose{2}{}, \cdots, \localizedPose{N}{}\}$, \ie~the \textcolor{blue}{\textbf{localized query frames}} poses with respect to the reference frame. Our approach does not need any prior geometry relative to the reference images (\eg~triangulated feature points), but it only relies on posed reference images to achieve this task.}
    \label{fig:objective}
\end{figure}

In this letter, we introduce \method (Map-free Augmented Reality Localization), a novel visual localization framework for \gls{ar} applications. 
Our approach uses a sequence of query images to create a local representation of the 3D structure around the current position of the user and then leverages this local representation to accurately estimate the relative pose to a given world-registered reference image. While the best-performing methods for map-free localization mainly employ techniques from Machine Learning \cite{arnold2022mapfree,barroso2024mickey,Leroy2024GroundingIM}, \method demonstrates that 
classical Computer Vision approaches
can also achieve excellent performance in the map-free case.
We extensively evaluate \method against several established methods for both map-free and map-based visual localization, demonstrating its \sota performance in robustness and accuracy among map-free approaches. 

While most of the existing literature shows functionality only on pre-recorded benchmark datasets, leaving the practicality of the approach open, we additionally demonstrate the real-world applicability of \method in a real-world scenario with a Magic Leap 2~\footnote{\url{https://www.magicleap.com/magic-leap-2}} \gls{ar} headset and Mapillary~\footnote{\url{https://www.mapillary.com}} reference images. To the best of our knowledge, this is the first real-world demonstration of a map-free localization approach with a commercially available \gls{ar} headset.

In summary, our key contributions are threefold:

\begin{itemize} \itemsep0em
    \item We introduce \method, a novel method for \textit{map-free localization}.
    \item We demonstrate that techniques from \sfm and \slam can also be successfully applied to the map-free localization problem.
    \item We extensively evaluate \method on benchmark datasets, attesting to its competitive performance compared to the current \sota in map-free localization.
    \item We conduct a real-world experiment, demonstrating the practical applicability of \method with a commercially available \gls{ar} headset.
\end{itemize}

%% file: 2-related_work.tex

\section{Related works}\label{sec:related}

Visual Localization is an established problem in the Computer Vision and Robotics community. Although several methods have been introduced over the years to efficiently solve this problem using the existing geometry of the reference environment~\cite{sarlin2022lamar}, our paper focuses on map-free scenarios~\cite{arnold2022mapfree}, which are particularly relevant for \gls{ar}~applications.

\mypar{Structure-Based Localization} Traditional structure-based visual localization methods~\cite{schoenberger2016mvs,Sattler2012ImprovingIL,sarlin2019hloc,sarlin2022lamar} estimate camera poses by first building a 3D map via \sfm \cite{schoenberger2016sfm}, and then finding correspondences between the map features and local image descriptors~\cite{Lowe2004DistinctiveIF,Bay2006Surf,Rublee2011ORB,detone18superpoint,Dusmanu2019D2NetAT,Revaud2019R2D2,Yi2016LIFT,sarlin20superglue,lindenberger2023lightglue}. Large-scale localization approaches tackle this problem in a hierarchical fashion, leveraging image retrieval~\cite{Arandjelovi2015NetVLAD,Revaud2019ApGem,Cao2020UnifyingDL,Rau2020ImageBoxOverlap,Tolias2013MatchKernels} to reduce the search space of feature matching, keeping the computational complexity of the solution relatively bounded. These methods are the typical choice of visual localization when a large number of images are available to construct a map. However, sparsity in the number of images can lead to a lack of viewpoint overlap, thus highly decreasing the quality of the map's structure. Additionally, building a map requires an explicit mapping session, which can, including post-processing, take multiple hours for medium-sized environments. This may impose a significant burden or not be possible at all, depending on the application. On the contrary, our method does not require the user to perform an explicit mapping session to achieve accurate localization results.

\mypar{Keypoint Matching + Depth Estimation} Given two images with enough overlap and known intrinsics, the relative transformation between them can be estimated through essential matrix decomposition~\cite{Hartley2004MVG,Nistr20045point,Hartley19978point} by making use of local 2D features. Still, this family of methods can only estimate the relative translation vector up to an unknown scale. If additional depth information is available, the problem can be solved using \gls{pnp} algorithms~\cite{Lee2013PnP} or 3D-3D registration techniques~\cite{Eggert1997Estimating3R,Besl1992AMF}. Recent advancements in deep learning resulted in several frameworks to estimate depth from a monocular image~\cite{Ranftl2020DPT1,Ranftl2021DPT2,Yang2024DA,Wang2024dust3r}, and the community has investigated their usage for map-free localization purposes~\cite{arnold2022mapfree}. However, the accuracy of the localization outcome is bounded by the accuracy of the depth estimation methods, which might struggle to have consistent depth predictions around corners and edges - \ie~typical parts of the images where 2D features are extracted. While these methods rely on monocular depth estimation to lift keypoints into 3D space, our method overcomes this limitation by leveraging known relative transformations between the query sequence of images to infer the 3D geometry of keypoints.

\mypar{\rpr} Recent deep learning approaches directly estimate the relative transformation between two images~\cite{Balntas2018RelocNetCM,Winkelbauer2021ExReNet,Chen2021WideBaselineRC,En2018RPNetAE,arnold2022mapfree}. These methods use a single network to extract encodings of both images and use them to regress an estimate of the relative transformation between the images. However, these methods cannot predict the quality confidence of the regressed pose, which is often useful to discard estimates when there is not enough overlap between the images, ultimately resulting in many outliers in the estimated poses. Additionally, these methods are often less accurate than matching-based approaches when image overlap is not a limiting factor.

\mypar{Differential RANSAC} RANSAC~\cite{Fischler1981RandomSC} is a robust method to estimate the parameters of a mathematical model. Differentiating RANSAC~\cite{Brachmann2016DSACD,brachmann2019ngransac} allows one to learn the relative pose between two images in an end-to-end manner. MicKey~\cite{barroso2024mickey} proposes to leverage differential RANSAC to directly learn the 3D-3D correspondences between two images end-to-end. Similarly to~\cite{roessle2023e2emultiviewmatching}, MicKey uses the relative pose between the two images as the only supervision signal. While this family of method gains some robustness over \rpr from leveraging RANSAC, accuracy is still limited compared to strucutre-based approaches.

\mypar{Image matching in 3D} Wang \etal propose  DUSt3R~\cite{Wang2024dust3r}, leveraging Vision Transformers~\cite{Kolesnikov2021ViT} (ViTs) to regress a dense scene representation from a pair of images in an end-to-end manner. The scene representation encapsulates a pointcloud together with its 2D-3D correspondences with the pair of images. The recently introduced MASt3R~\cite{Leroy2024GroundingIM} builds upon DUSt3R to add a new head that outputs dense local features, which is trained with a matching loss. This approach significantly improves the absolute pose estimation capabilities of DUSt3R and achieves \sota results in map-free localization. However, these methods still lack the precision achieved by structure-based approaches.

\mypar{Image Sequence Based Localization}Sequential methods for visual localization~\cite{Stenborg2020UsingIS,Gim2015MinimalSolutions,Milford2012SeqSLAMVR} leverage the continuity in image data to enhance localization robustness. These approaches have shown to be particularly beneficial under challenging conditions, such as seasonal or lighting changes \cite{Milford2012SeqSLAMVR}. Additionally, odometry measurements can be combined with the image sequences to further increase the accuracy and robustness of the localization process~\cite{Stenborg2020UsingIS}. Recently, LaMAR~\cite{sarlin2022lamar} has presented a comprehensive evaluation of sequence-based localization, demonstrating how longer query sequences can positively affect localization recall of AR devices. However, these methods either target improvements for structure-based localization methods, or are part of \slam systems, and, therefore, are not well-suited for map-free localization. In our method, we leverage the additional information provided by image sequences and adapt it to address the map-free localization challenge.

%% file: 3-background.tex
\section{Problem statement} 
\label{sec:problem}

\mypar{Notation} In the remainder of this document, $\pose{i}{j} \in \sethree$ denotes the 6-\dof transformation from coordinate frame $j$ to frame $i$. $\pose{i}{j} =  \begin{bmatrix}
    \rot{i}{j} \, | \, \translation{i}{j}
\end{bmatrix}$ is parametrized by a rotation $\rot{i}{j} \in \sothree$ and a translation $\translation{i}{j} \in \mathbb{R}^3$. We denote $q$ and $r$ as the query and reference frames, respectively. $\keypoint{q}{i}{k}$ denotes the image coordinate of the $k$-th 2D keypoint of image $\queryImage{i}$ corresponding to the query sequence; similarly, $\keypoint{r}{j}{k}$ denotes a keypoint of the reference image $\refImage{j}$. $\liftedKeypoint{q}{i}{k}$ denotes the triangulated 3D pose in the query frame of the $k$-th keypoint of image $\queryImage{i}$ corresponding to the query sequence.

\mypar{Objective} Given a sequence of $N$ posed query images $\querySequence = \{\queryImage{1} \; \cdots \; \queryImage{N}\}$ with corresponding poses $\queryPoses = \{\queryPose{1} \; \cdots \; \queryPose{N}\}$ in the query frame $q$, and a set of $M$ posed reference images $\refSequence = \{\refImage{1} \; \cdots \; \refImage{M}\}$ with corresponding poses $\refPoses = \{\refPose{1} \; \cdots \; \refPose{M}\}$ in the reference frame $r$, our goal is to find the query frame poses with respect to the reference frame - \ie~$\{\localizedPose{1}{} \; \cdots \; \localizedPose{N}{}\}$. Note that, in the extent of this paper, we assume to have no prior geometry associated with the reference images, meaning that, no explicit mapping stage via \slam or \sfm has to be performed by the user - reference images and associated poses in the reference frame are sufficient to use \method. We present a visual representation of the problem statement in~\figref{fig:objective}.

%% file: 4-method.tex

\section{Method}\label{sec:method}

 We introduce \method, a novel pipeline for map-free localization. Our approach takes as input a sequence of monocular images or rigs with multiple images posed in the query reference frame, together with one or more reference images located in a different (global) reference frame. The output is the set of 6DoF poses relative to each individual query image expressed in the global reference frame. Note that no prior geometry is required for \method to work. Since \gls{ar} devices are required to track their motion through the environment to display world-registered content, the relative poses of the query frames can be easily obtained from the onboard \gls{vio} or \slam system of the \gls{ar} device, such as \textit{ARKit}, \textit{ARCore}, or Magic Leap's \textit{Head Tracking}. Therefore, we assume that these poses are given, and estimating them is outside the scope of this method.

\begin{figure}[t]
    \centering
    \includegraphics[width=1\columnwidth]{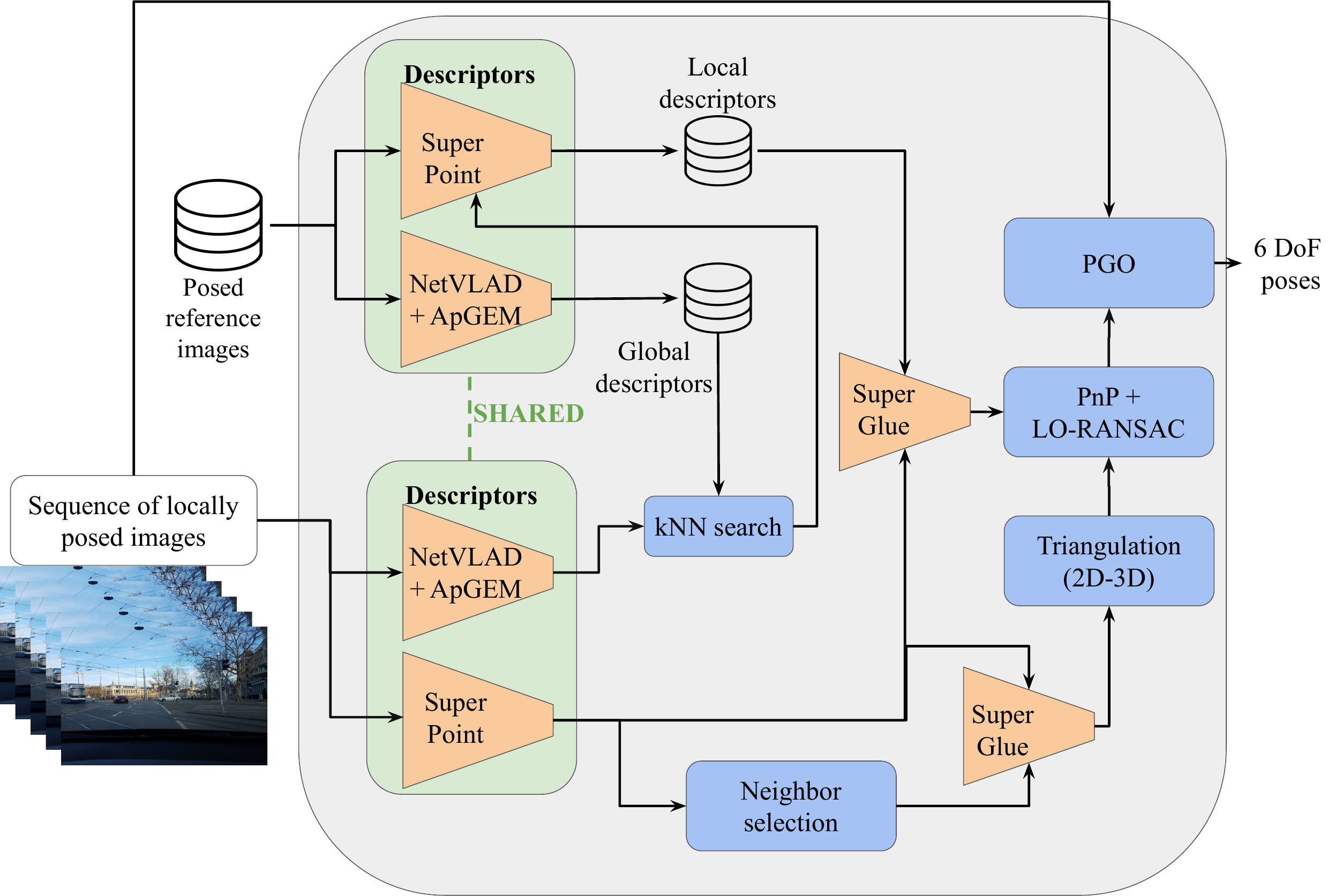}
    \caption{The query frames are localized using only reference images from the database. \method uses NetVLAD~\cite{Arandjelovi2015NetVLAD} + APGeM~\cite{Revaud2019ApGem} for image retrieval, while SuperPoint~\cite{detone18superpoint} and SuperGlue~\cite{sarlin20superglue} are used to extract and match local descriptors. We triangulate query 3D points by leveraging the prior poses of the query frames, allowing us to perform pose estimation via the \gls{pnp} algorithm. The final set of query poses is refined via \gls{pgo}.}
    \label{fig:overall}
\end{figure}

In the remainder of this section, we outline the main individual components of our approach, summarized in~\figref{fig:overall}. We start by performing feature extraction and candidate selection in~\secref{subsec:candidate_selection}, which is followed by local feature extraction and matching. We then lift the query 2D features by performing intra-sequence triangulation in~\secref{subsec:query_sequence_triangulation}. Finally, we make use of the triangulated points to perform pose estimation and refine our pose estimates through relative pose graph optimization in~\secref{subsec:pose_estimation}.

\subsection{Candidate Selection and Matching}
\label{subsec:candidate_selection}
Following LaMAR~\cite{sarlin2022lamar}, we extract a fusion~\cite{Humenberger2020Fusion} of the NetVLAD~\cite{Arandjelovi2015NetVLAD} and APGeM~\cite{Revaud2019ApGem} descriptors for all query and reference images. We make use of the extracted descriptors to perform image retrieval on the reference images. Thus, for each query image $\queryImage{i}$ we select the closest reference images $K$ $\{\refImage{j}\}_{1\leq j \leq K}$, referred to as \emph{candidates}.

Subsequently, we use SuperPoint~\cite{detone18superpoint} to extract local features from the query sequence and the selected candidate images from the reference set. Then, we use SuperGlue~\cite{sarlin20superglue} to perform feature matching and obtain a set of 2D-2D correspondences. Formally, for each query image $\queryImage{i}$ and its selected candidate reference images $\{\refImage{j}\}_{1\leq j \leq K}$, we obtain the 2D-2D correspondences $\{\keypoint{q}{i}{k}\, \keypoint{r}{j}{k}\}_k$.

\subsection{Intra-sequence Query Keypoints Triangulation}
\label{subsec:query_sequence_triangulation}

Common visual localization algorithms extract 2D-3D correspondences by lifting the keypoints of reference images, either using previously built geometry representations of the reference environment~\cite{sarlin2022lamar,sarlin2019hloc,schoenberger2016mvs,schoenberger2016sfm} - \eg~sparse or dense point clouds - or leveraging depth information from additional sensors or extracted from the monocular image~\cite{arnold2022mapfree}. In this work, we propose to make use of the known relative transformations between frames of the query sequence to triangulate 2D keypoints and lift them to 3D. These 3D points are used to construct 3D-2D correspondences between the query and reference frames. In the remainder, we describe the proposed steps to triangulate query keypoints.

\mypar{Neighbor selection} To perform intra-sequence triangulation on the query images, we first select a neighbor image (from within the query sequence) for each query image. Since we need a minimum of parallax between a query frame and its selected neighbor, we propose to use~\algref{alg:neighbor_selection}, which performs a simple distance check on the rotational and translational displacement between the two. Thus, for query image $\queryImage{i}$, we select its neighbor as the first frame in the sequence whose relative pose compared to the query frame is larger than a user-specified threshold ($\bt_{min}, \theta_{min}$). For each query frame, if a neighbor is not found due to a lack of distance between query frames, the pose estimation for the respective query frame is skipped.

 \begin{algorithm}\caption{Neighbor Selection}\label{alg:neighbor_selection}
    \begin{algorithmic}[1]
      \State \textbf{Input:} $\queryPoses$, $\bt_{min}, \theta_{min}$
      \State $\neighborSequence \gets []$
      \For {$i = 0, \hdots, N-1$}
        \For {$j = i+1, \hdots, N$}
        \State $\pose{i}{j}^{e} \gets (\pose{q}{i})^{-1} \pose{q}{j}$
        \State $\quaternion{i}{j}^{e} \gets \text{rot2quat}(\rot{i}{j}^{e})$
        \State $\theta^{e} \gets \text{atan2}(||\quaternion{i}{j}^{e}[x, y, z]||_2, ||\quaternion{i}{j}^{e}[w]||_1)$
        \If{$||\translation{i}{j}||_2 \geq \bt_{min} \quad || \quad \theta^{e} \geq \theta_{min}$}
            \State $\neighborSequence[i] \gets j$
            \State \textbf{break}
        \EndIf
        \EndFor
      \EndFor 
    \State \Return $\neighborSequence$
    \end{algorithmic}
  \end{algorithm}

\mypar{Feature matching and triangulation} Next, we match local features for each query sequence image and their corresponding neighbor. Thus, for each query image $\queryImage{i}$ and its corresponding neighbor, denoted as $\neighborSequence[i]$, we obtain a set of 2D-2D correspondences $\{\keypoint{q}{\neighborSequence[i]}{k}\, \keypoint{q}{i}{k}\}_k$. Consecutively, we make use of the matches to triangulate 3D points in the query frame using their poses $\queryPose{i}$ and $\queryPose{\neighborSequence[i]}$, obtaining $\{\keypoint{q}{i}{k}\, \liftedKeypoint{q}{i}{k}\}_k$.

\mypar{3D-2D correspondences} Finally, we extract 3D-2D correspondences between the reference and query frames by selecting the 2D keypoints present in both the query to reference images 2D-2D matches $\{\keypoint{q}{i}{k}\, \keypoint{r}{j}{k}\}_k$ and the query to neighbor 2D-2D matches $\{\keypoint{q}{\neighborSequence[i]}{k}\, \keypoint{q}{i}{k}\}_k$. Thus, for each query and retrieved reference images, we obtain a set of 3D-2D correspondences: $\correspondences_i = \bigcup_{j=1}^{K}\{\liftedKeypoint{q}{i}{k} \, \keypoint{r}{j}{k}\}_k$. Note that, in our approach, the 3D points are computed in the local query frame. Therefore, we do not make use of or extract any structure in the reference frame - we simply use the posed reference images.

\subsection{Pose Estimation and Pose Graph Optimization (PGO)}
\label{subsec:pose_estimation}

Having computed the 3D-2D correspondences $\correspondences_i$, 
we employ a standard PnP~\cite{Lee2013PnP} algorithm to obtain an initial estimate of the query poses with respect to the reference frame, namely $\localizedPose{i}{,init}$. 
To this end, we employ the pose estimation module available in
COLMAP~\cite{schoenberger2016sfm}, which makes use of a LO-RANSAC~\cite{Chum2003RANSAC} stage and a non-linear refinement. 
Note that we discard all poses that contain a number of inliers lower than a user-specified threshold. 
This can happen due to a lack of texture, not enough overlap, difficult lighting conditions, or domain changes - \eg~weather and seasons. Typically, given a set of 2D-3D correspondences, PnP computes the pose of the camera with respect to the 3D points' origin. Therefore, with our formulation, the result of PnP is the transformation from the reference frame to the local query sequence frame $\pose{r}{q}$. By inverting this transformation, $\pose{q}{r} = \pose{r}{q}^{-1}$, we can compute the desired poses $\localizedPose{i}{,init} = \pose{q}{r}\queryPoses$.

As a final step, we refine our pose estimates by leveraging relative Pose Graph Optimization (PGO)~\cite{grisetti2020least} between the estimated poses and the localization queries. 
In detail, we compute the optimal set of poses $\localizedPose{i}{, PGO}$ that minimize the following cost function:
\begin{equation*}
    \min_{\localizedPose{i}{, PGO}}{\sum_{i=1}^{N-1}} \be_i^T \Sigma^{-1} \be_i,
    \label{eq:pgo-loss-generic}
\end{equation*}
where each cost factor $i$ is an $\mathrm{SE}(3)$ constraint defined as
\begin{equation}
    \be_i = \left((\localizedPose{i+1}{, PGO})^{-1} \localizedPose{i}{, PGO}\right) \boxminus \left((\queryPose{i+1})^{-1}\queryPose{i}\right),
    \label{eq:pgo-error}
\end{equation}
and $\Sigma \in \bbR^{6\times6}$ is the covariance matrix of the localization queries provided by the on-device \gls{vio} or \slam system.
The $\boxminus$ operator defined in~\eqref{eq:pgo-error} is a non-linear mapping $\mathrm{SE}(3) \rightarrow \mathbb{R}^n$ that computes the local distance between the two poses in the manifold~\cite{lee2003introduction, Sol2018AML}. Furthermore, we employ a robust loss function to handle eventual outliers in the optimization~\cite{mactavish2015all, Tukey1974, Huber1964RobustEO}.

With this operation, we use the information stored in the query poses, computed by the on-device \slam or \gls{vio} system, to constrain the relative motion of subsequent localization estimates. We initialize the optimizable poses $\localizedPose{i}{, PGO}$ to $\localizedPose{i}{, init}$, which have been estimated previously with PnP and LO-RANSAC. Additionally, to avoid an ill-posed formulation, we fix the pose $\localizedPose{i}{, PGO}$ with the most number of inliers.

%% file: 5-experimental_results.tex

\section{Experiments}\label{sec:results}

In this section, we provide quantitative and qualitative results obtained with our \method pipeline. We compare \method with the most recent \sota methods in the field for map-free localization, and demonstrate our proposed approach in a real-world \gls{ar} setting, using a Magic Leap ML2 device and the Mapillary crowd-sourced database to precisely register digital content with the physical world. To the best of our knowledge, this is the first time that a map-free visual-localization pipeline has been integrated into a commercially available \gls{ar} headset.

\mypar{Implementation Details}  We integrate our proposed method into the localization framework provided by LaMAR~\cite{sarlin2022lamar}~\footnote{\url{https://github.com/microsoft/lamar-benchmark}}. To select a neighbor for each query frame, we set the minimum distance thresholds used in \algref{alg:neighbor_selection} to $\bt_{min}=0.3m$ and $\theta_{min}=10\degree$. The default configurations~\cite{sarlin2022lamar}  for NetVLAD~\cite{Arandjelovi2015NetVLAD}, ApGEM~\cite{Revaud2019ApGem}, SuperPoint~\cite{detone18superpoint} and SuperGlue~\cite{sarlin20superglue} are used. When triangulating 3D points from 2D-2D correspondences between a query and its neighbor, we allow for a maximum reprojection error of 3 pixels. Additionally, we apply the same maximum reprojection error during pose estimation (\secref{subsec:pose_estimation}). Finally, we set the sequence query batch $N=10$. The queries in a sequence are consecutive (\ie~timestamp ordered) in all datasets and when testing with the AR device. Each batch is processed individually and does not incorporate localization results from previous batches into the process.

\mypar{Datasets} \method is evaluated in two datasets: the Niantic Map-Free Relocalization Dataset~\cite{arnold2022mapfree} and the LaMAR dataset~\cite{sarlin2022lamar}. The Niantic Map-Free Relocalization Dataset is the standard benchmark of map-free localization methods. It consists of 655 outdoor scenes containing small places of interest. The dataset is split into 460 training scenes, 65 validation scenes, and 130 test scenes. Each scene contains a single reference image and a set of approximately 500 query images. In this work, we use the validation dataset only to generate localization queries, since the ground truth poses of the query images with respect to the reference image are provided. The LaMAR dataset is a large-scale dataset captured using \gls{ar} devices, and, hence, the results provide insights into how well each method performs in real-world \gls{ar} settings. Specifically, three different sessions (CAB, HBE and LIN) are captured, containing both indoor and outdoor environments. Each session includes a mapping sequence, which we use as a reference, and two validation query sequences (captured with a Microsoft HoloLens 2~\footnote{\url{https://www.microsoft.com/en-us/hololens}} and hand-held Apple iPhone/iPad) which include ground truth poses with respect to the mapping sequence. 

\mypar{Baselines} We compare \method against \sota methods in map-free localization: \textbf{i)} MicKey~\cite{barroso2024mickey}, \textbf{ii)} Relative Pose Regression parametrized with a 6D parametrization for the rotation (referred to as RPR [$R(6D) + t$])~\cite{arnold2022mapfree,Zhou2018OnTC}, \textbf{iii)} feature matching (with SuperPoint and SuperGlue) with essential matrix estimation and scale from monocular depth from~\cite{arnold2022mapfree} (shortened as E+D), and \textbf{iv)} MASt3R~\cite{Leroy2024GroundingIM}. 
Furthermore, we include comparisons with the LaMAR~\cite{sarlin2022lamar} framework, as representative of structure-based hierarchical localization. The comparison with the LaMAR pipeline has been performed using the companion dataset from the same authors, providing insights of how well the proposed approach performs compared to the current gold standard in visual localization. Note that all baselines that we compare against~\cite{barroso2024mickey,arnold2022mapfree,Leroy2024GroundingIM,sarlin2022lamar} explicitly design their method to perform \textit{metric} pose estimation, and present their evaluations in $\sethree$~errors. Consistently, the \textit{map-free} localization field~\cite{arnold2022mapfree} was explicitly introduced to estimate metric poses from pairs of query-reference images.

\mypar{Metrics} Following~\cite{sarlin2022lamar,arnold2022mapfree}, we evaluate \method and the baselines using the median Absolute Pose Error (APE), both in translation and rotation. Additionally, we evaluate the recall at different translation and rotation error thresholds. The recall is computed as the fraction of correctly localized queries within an error radius, which is determined by the translation and rotation thresholds. For example, 
$\mathrm{recall}@(0.5m, 1\degree)$ 
indicates the fraction of localized queries with an error margin of $0.5 m$ in translation and $1 \degree$ in rotation. Furthermore, we report the time required to localize a query image sequence.

\subsection{Niantic Map-Free Relocalization Dataset}\label{subsec:niantic_mapfree}

In our comparisons, we include all map-free baselines. Note that in this dataset, the number of candidates used in our method $K$ is always 1, since there is only one reference image per scene. Therefore, LaMAR is not evaluated in this dataset since structure cannot be extracted from a single reference image. \tabref{tab:quantitative_mapfree} presents the main results on the Niantic Map-Free dataset. As shown, our method achieves \sota results in terms of median APE. The table additionally shows the percentage of localized queries. Methods that use matching - \ie~ours and E+D - take into account the number of inliers to reject outlier queries when this number is low. 
For this reason, the percentage of localized queries in MicKey and RPR [$R(6D) + t$] is higher compared to the matching-based methods, however, at the cost of significantly reduced average accuracy.

\begin{figure}[h]
        \centering
        \includegraphics[width=\columnwidth]{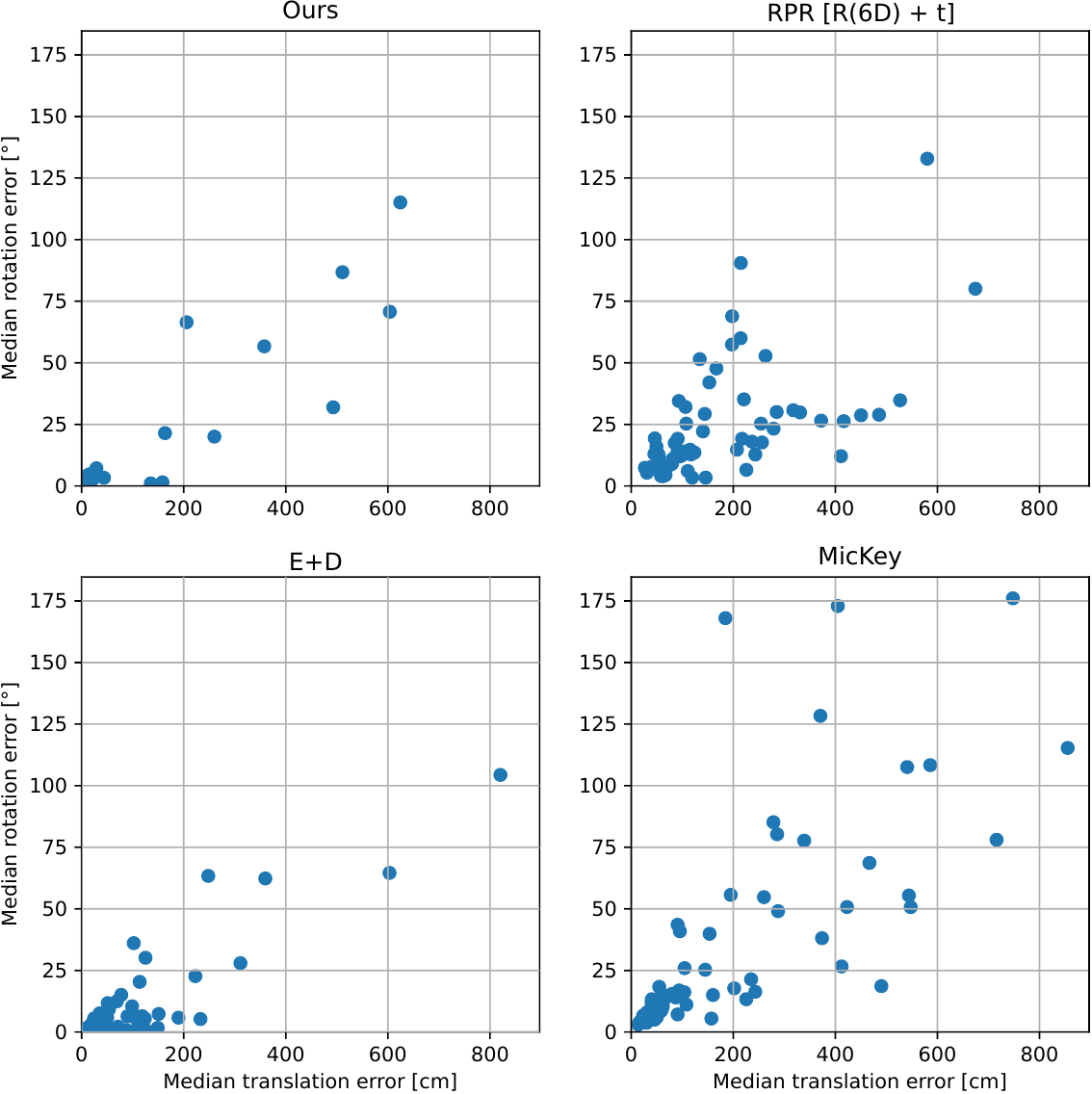}
        \caption{Median rotation error against median translation error for each scene of the \textbf{Niantic Map-Free dataset}. The error distribution of our approach is concentrated in the bottom-left corner, with the center of mass in a low-error range.}
        \label{fig:mapfree_scatter}
\end{figure}

\begin{figure}[h]
        \centering
        \begin{tabular}{cc}
             \includegraphics[width=.5\columnwidth]{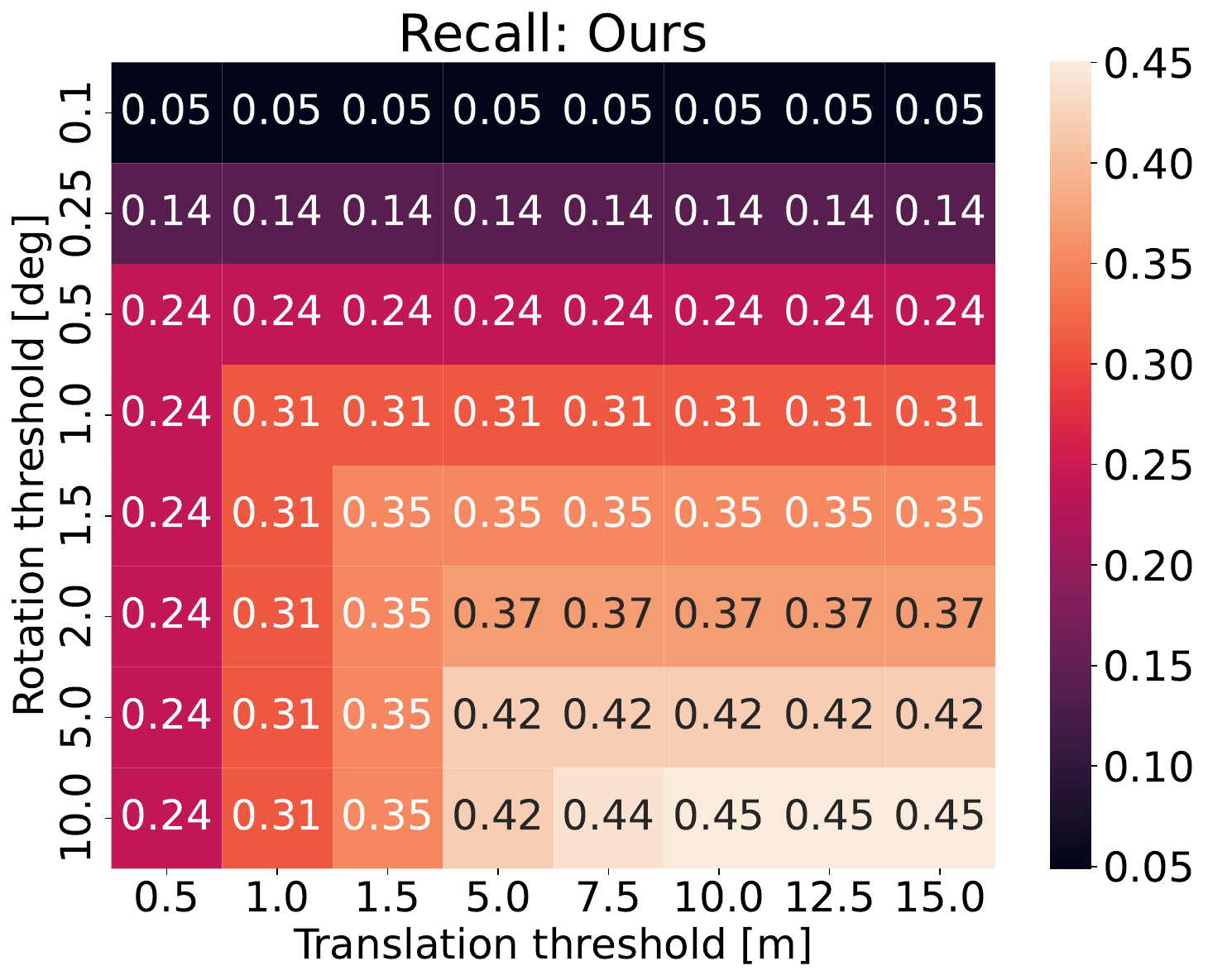} & \includegraphics[width=.5\columnwidth]{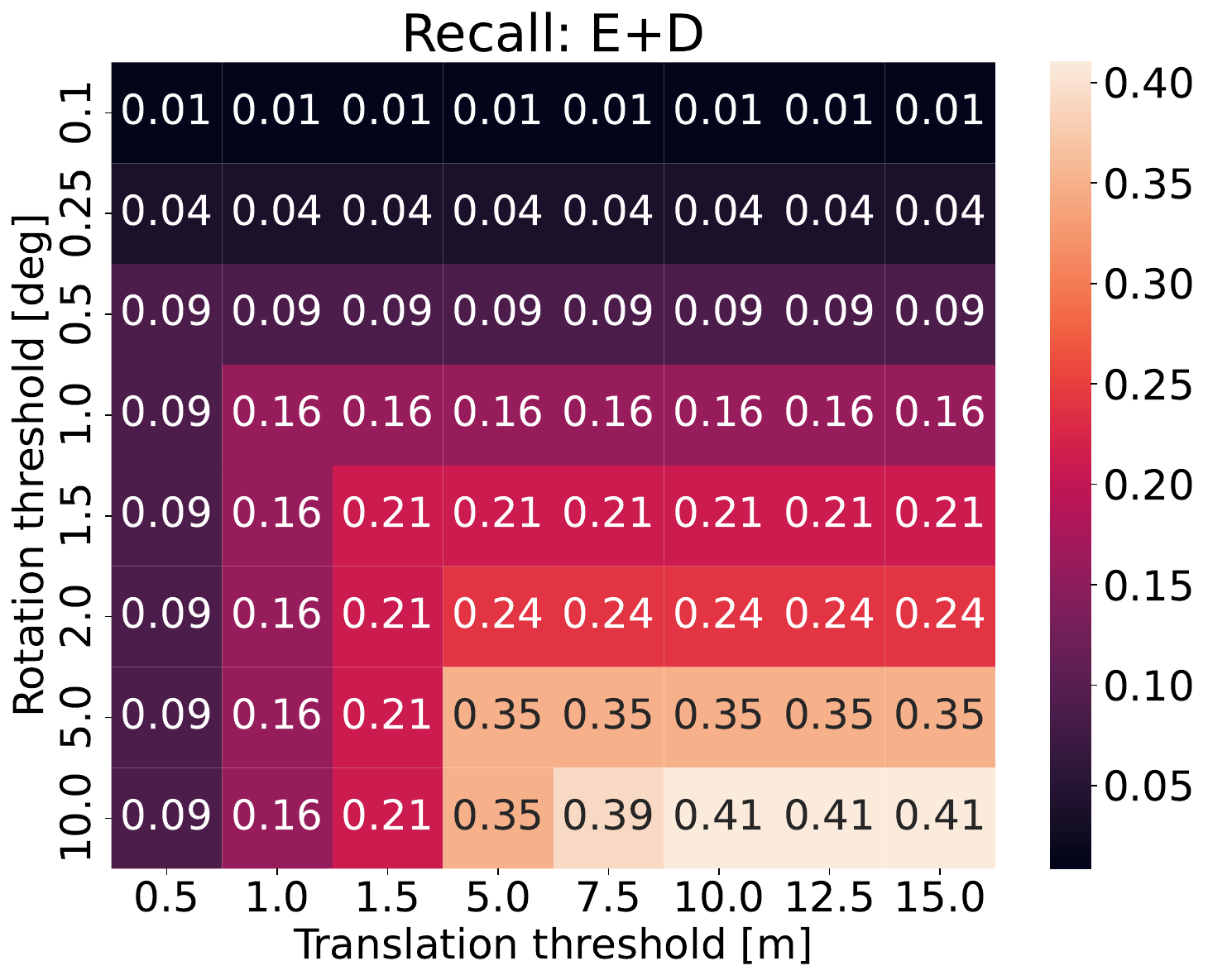} \\
             \includegraphics[width=.5\columnwidth]{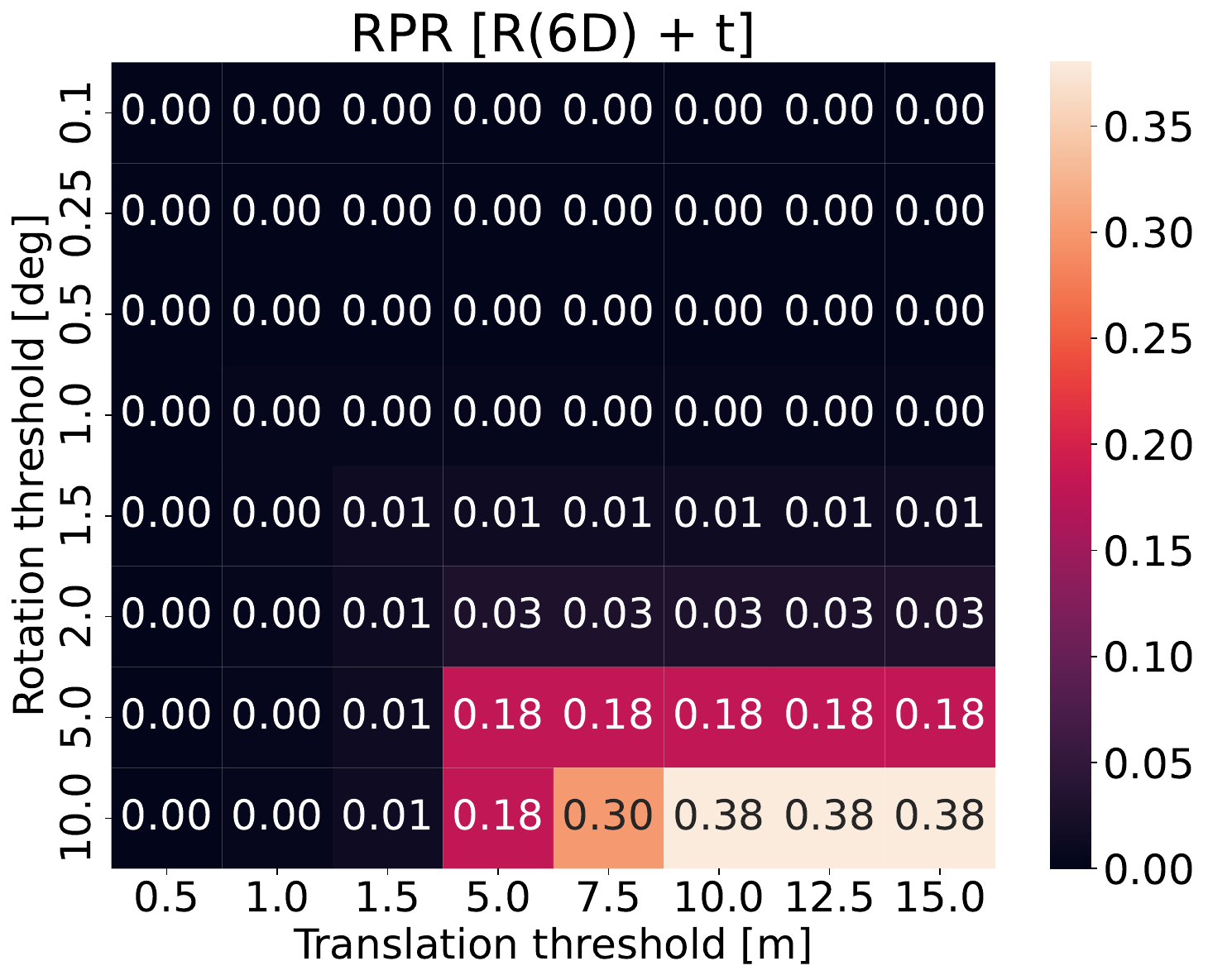} &{\includegraphics[width=.5\columnwidth]{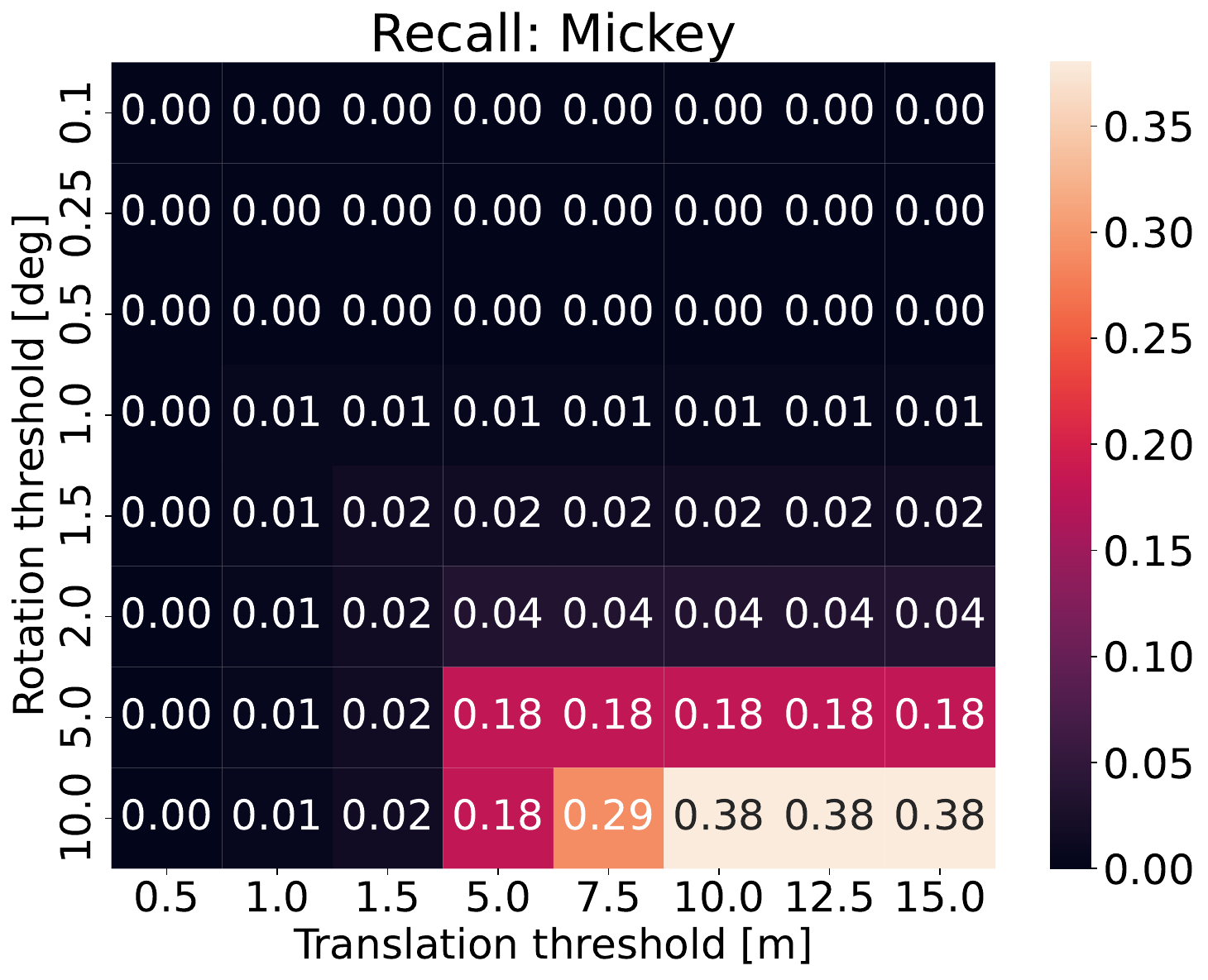}}
        \end{tabular}
        \caption{Recall in \textbf{Niantic Map-Free dataset} at different APE thresholds. Best viewed when zoomed in.}
    \label{fig:mapfree_recall}
\end{figure}

\figref{fig:mapfree_scatter} presents the APE values evaluated on each scene of the dataset. Our method concentrates most of the samples at low values, while other baselines present a larger spread, which attests to the robust performance of our \method. We additionally showcase the recall at different APE thresholds in~\figref{fig:mapfree_recall}. As visible from the figures,  \method achieves a higher recall than all baselines at any APE. An ideal method would have a recall of $1.0$ at any APE threshold, therefore, we aim to achieve the maximum possible recall at any APE. Typically, larger APE thresholds allow higher recall, since the error margin is higher. As visible from the figures,  \method achieves a higher recall than all baselines at any APE threshold.

\setlength{\tabcolsep}{3pt}
\begin{table}[h]
    \centering
    \begin{tabular}{l|ccc}
        \toprule
         & \textbf{Median} & \textbf{Average} & \textbf{\% localized}\\
         & \textbf{APE} & \textbf{query time} & \textbf{queries}\\
         \midrule
         MicKey~\cite{barroso2024mickey} & 94.7 cm, 14.52\degree & 19.68 s & \boldgreen{100} \\ 
         RPR [$R(6D) + t$]~\cite{arnold2022mapfree} & 123 cm, 15.31\degree  & \boldblue{0.04} s & \boldgreen{100}\\ 
         E+D~\cite{arnold2022mapfree} & 57.5 cm, 2.28\degree &\boldgreen{0.03 s} & 51.76\\
         MASt3R~\cite{Leroy2024GroundingIM}~\tablefootnote{Results for MAST3R on the Niantic Map-Free dataset as reported in ~\cite{Leroy2024GroundingIM}.} & \boldblue{35 cm, 2.2\degree} & N/A & N/A \\
         \midrule
         \textbf{\method (ours)} & \boldgreen{5.55 cm, 0.80\degree} & 0.33 s & \boldblue{59.27}\\
        \bottomrule
    \end{tabular}
    \caption{We present a comparison of the median APE achieved by our method and the baselines using the \textbf{Niantic Map-Free dataset}. We additionally present the time per query achieved by each method and the percentage of localized queries. Our method clearly outperforms the baselines in terms of APE. \boldgreen{Best result}. \boldblue{Second best result}.}
    \label{tab:quantitative_mapfree}
\end{table}

\setlength{\tabcolsep}{1pt}

\subsection{LaMAR Dataset}\label{subsec:lamar_benchmark}

\setlength{\tabcolsep}{4.9pt}
\begin{table*}[t]
    \centering
    \begin{tabular}{l|ccc|ccc|ccc}
        \toprule
         & \multicolumn{3}{c|}{\textbf{Median APE}} & \multicolumn{3}{c|}{\textbf{Avg. Query Time}} & \multicolumn{3}{c}{\textbf{\% Localized}}\\
         & CAB & HGE & LIN & CAB & HGE & LIN & CAB & HGE & LIN\\
         \midrule
         MicKey~\cite{barroso2024mickey} & 427 cm, 25.29\degree & 459 cm, 19.75\degree & 255 cm, 17.62\degree & 3.43 s& 3.96 s & 13.92 s & \boldgreen{93.88} & \boldgreen{95.08} & \boldgreen{96.09} \\
         LaMAR ($K=10$)~\cite{sarlin2022lamar} & \boldgreen{3.25 cm, 0.30\degree} & \boldgreen{4.64 cm, 0.26\degree} & \boldgreen{1.96 cm, 0.17\degree} & 2.89 s & 5.83 s & 9.15 s & \boldblue{83.81} & 86.45 & 85.35\\
         LaMAR ($K=1$)~\cite{sarlin2022lamar} & \boldblue{3.58 cm, 0.34\degree} & \boldblue{6.29 cm, 0.30\degree} & \boldblue{2.17 cm, 0.18\degree} & \boldblue{0.52 s} & \boldblue{0.86 s} & \boldblue{1.28 s} & 54.11 & 66.89 & 72.65\\
         \midrule
         \textbf{\method} $\boldsymbol{(K=10)}$ \textbf{(Ours)} & 7.44 cm, 0.52\degree & 14.06 cm, 0.45\degree & 5.44 cm, 0.27\degree & 1.96 s & 1.98 s & 1.60s & 66.83 & \boldblue{93.91} & \boldblue{88.98}\\
         \textbf{\method} $\boldsymbol{(K=1)}$ \textbf{(Ours)} & 8.86 cm, 0.57\degree & 18.1 cm, 0.50\degree & 6.51 cm, 0.37\degree & \boldgreen{0.21 s} & \boldgreen{0.29 s} & \boldgreen{0.27 s} & 57.15 & 86.32 & 80.56\\
        \bottomrule
    \end{tabular}
    \caption{We present a comparison of the median APE achieved by our method and the baselines using the \textbf{LaMAR dataset}. We additionally present the time per query achieved by each method and the percentage of localized queries. Our pipeline achieves significantly better results compared to MicKey, producing metrics in the same range as LaMAR without the need for an explicit \gls{sfm} session. \boldgreen{Best result}. \boldblue{Second best result}.}
    \label{tab:quantitative_lamar}
\end{table*}

\method is compared against MicKey and LaMAR in the LaMAR dataset. Here we use a dataset with reference poses generated by \gls{sfm}, since this is required for LaMAR, however, the map-free approaches alone would not need such data, as shown in section \ref{subsec:niantic_mapfree}. We do not evaluate RPR [$R(6D) + t$] and E+D on this dataset, since retraining is required for both the regression and monocular depth estimation models. To have a fair evaluation, we present the results of LaMAR and our method using $K=1$ and $K=10$ retrieved reference image candidates since MicKey cannot use more than one retrieved candidate for pose estimation. Note that we use the same image retrieval method to select $K$ candidates for all methods, \ie~a fusion~\cite{Humenberger2020Fusion} of the NetVLAD~\cite{Arandjelovi2015NetVLAD} and APGeM~\cite{Revaud2019ApGem} descriptors. We present quantitative results in~\tabref{tab:quantitative_lamar}. Our method produces significantly more accurate localization estimates compared to the other map-free pipelines evaluated on this dataset. Furthermore, the metrics obtained are in a similar range with the well-established LaMAR pipeline, without the need for an explicit \gls{sfm}-based mapping session. Note that LaMAR and \method are more conservative when localizing a query compared to MicKey, since we restrict the minimum number of inliers during pose estimation. This is reflected in a lower percentage of localized queries for both LaMAR and \method, but also in significantly higher accuracy. 

\begin{figure}[h]
        \centering
        \begin{tabular}{cc}
             \includegraphics[width=.5\columnwidth]{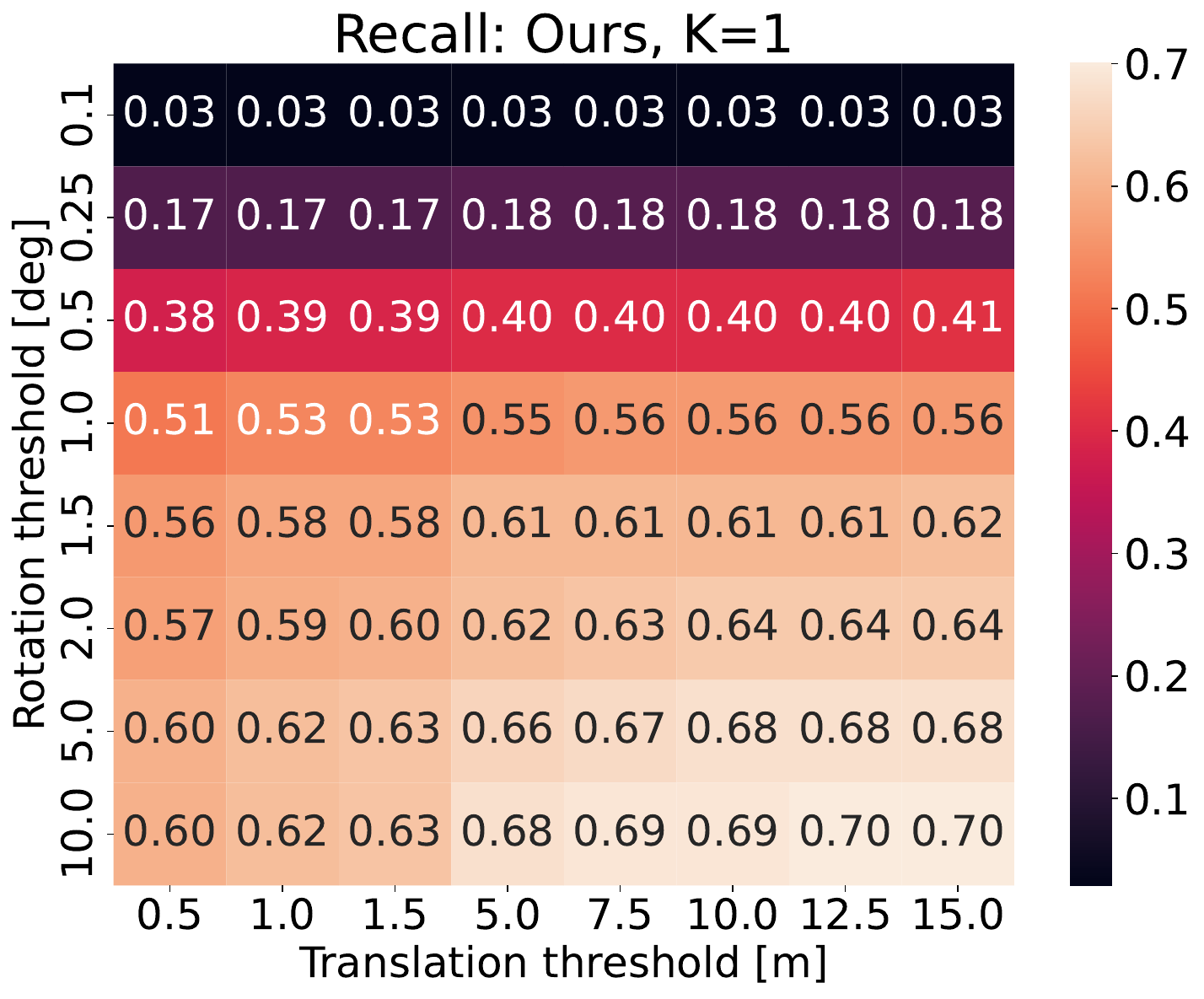} &
             \includegraphics[width=.5\columnwidth]{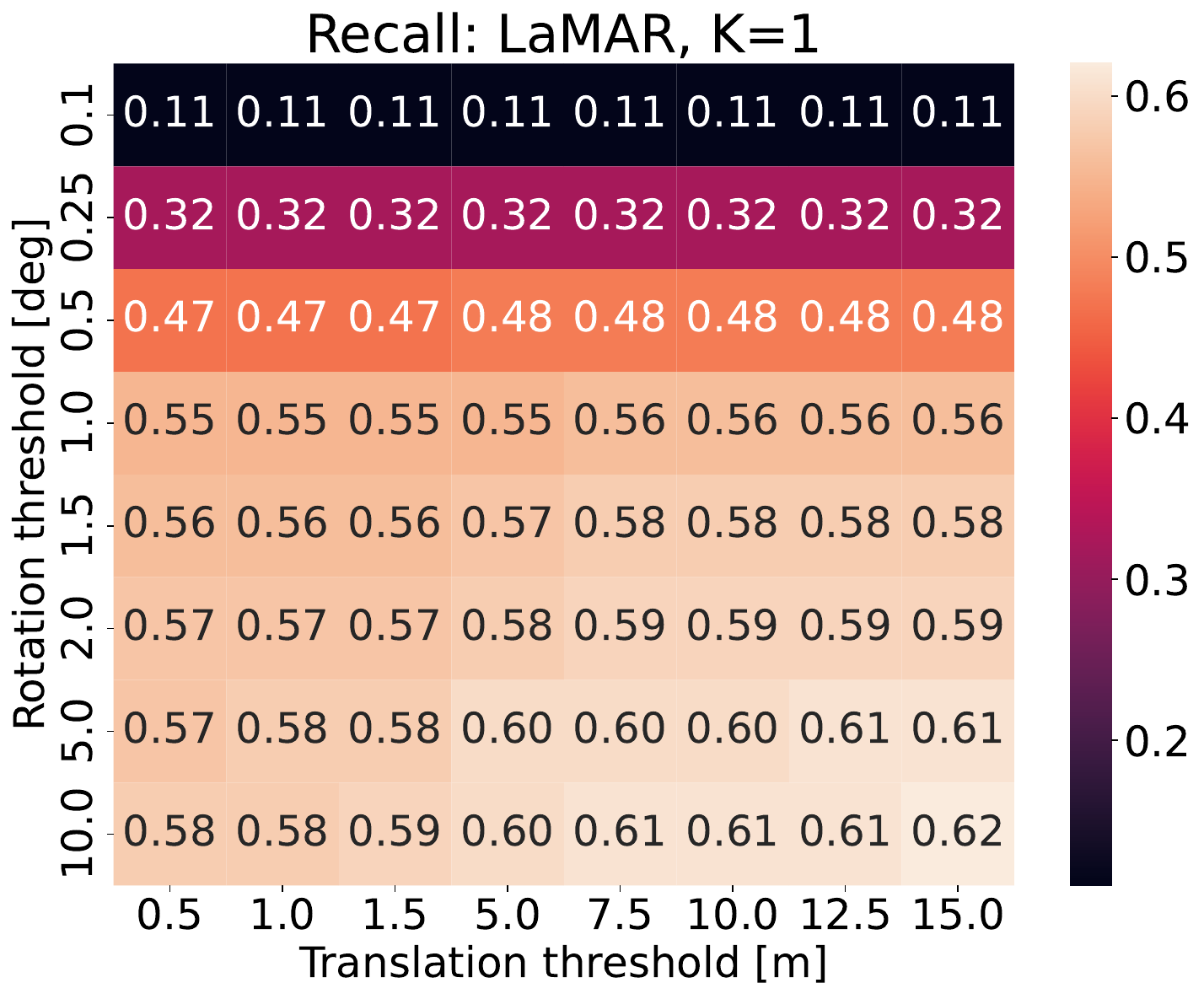} \\
             \multicolumn{2}{c}{\includegraphics[width=.5\columnwidth]{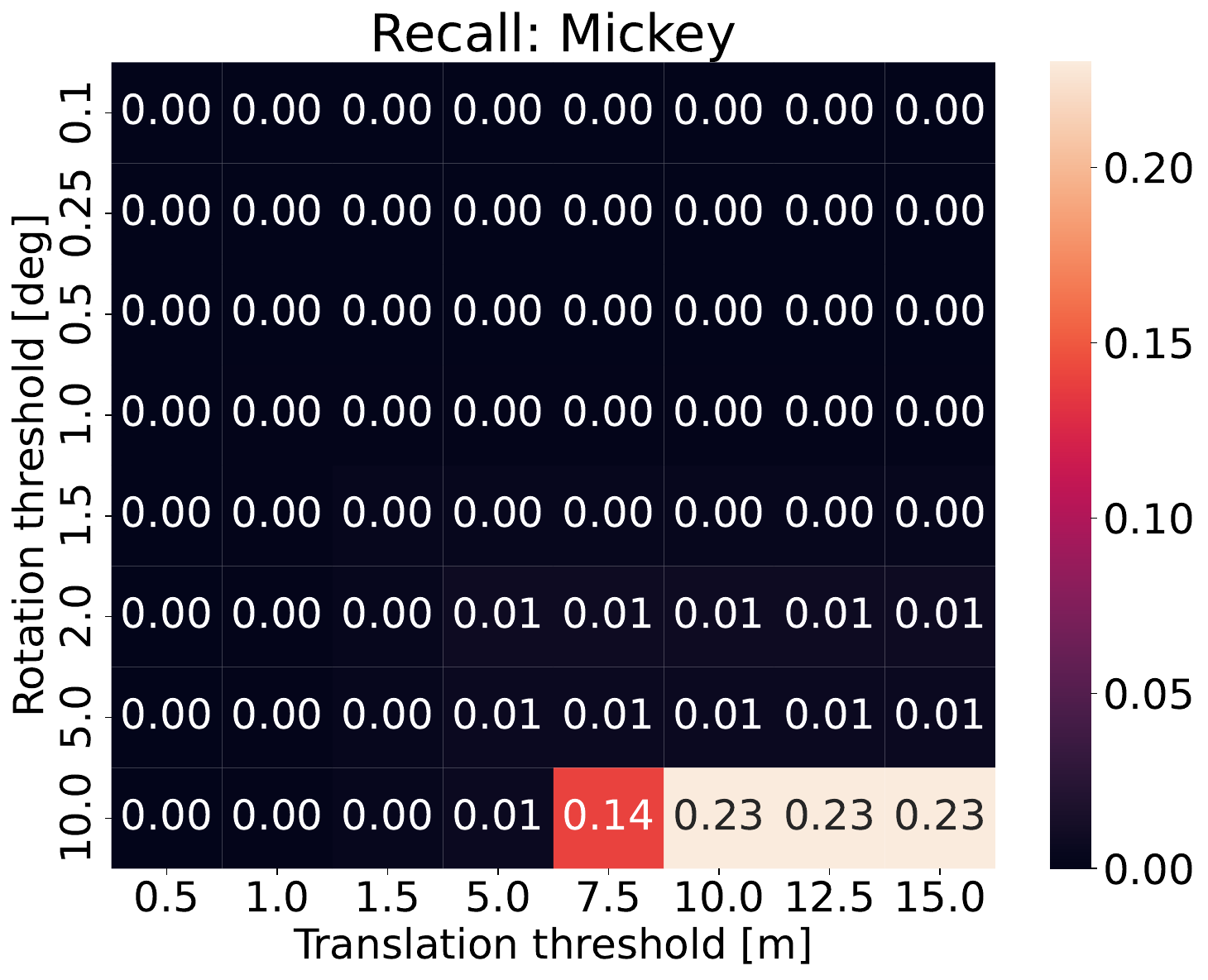}} \\
        \end{tabular}
        \caption{Recall in \textbf{LaMAR benchmark} at different APE thresholds. Best viewed when zoomed in.}
    \label{fig:lamar_recall}
\end{figure}

Furthermore, we report in~\figref{fig:lamar_recall} the recall of the three methods, evaluated at different pose error thresholds. 
\method significantly outperforms the other map-free methods and exhibits metrics in a similar range to the LaMAR framework, despite not using any prior structure at localization time.

\setlength{\tabcolsep}{1pt}

\subsection{AR Device Testing}\label{subsec:testing}

\mypar{Setup} 
To demonstrate the practicality of our approach in \gls{ar} applications, we use \method together with a Magic Leap 2 (ML2) device in a live, real-world experiment.  
The ML2 is equipped with a sensor rig that includes three cameras, specifically designed to accommodate Computer Vision algorithms, and has its own proprietary perception stack onboard. 
We collect ML2 \emph{rig-frames} (a frame combining one image per camera) at sensor rate as query images, while pose estimates for each frame are provided by the native \slam system of the ML2 and therefore expressed in the device reference frame $q$.
Each \emph{localization query} is composed of a set of $N=10$ posed rig-frames.

The images used as a reference for localization are taken from the public Mapillary database.
Mapillary provides geolocation information about each image - \ie~latitude, longitude and altitude - and it also exposes intrinsic parameters of the camera setup.
These Mapillary images and their geolocation are used as the reference sequence $\refSequence$. 
Our goal is to capture sequences of images with an ML2 and use them to localize the \gls{ar} device with respect to a global frame of reference using Mapillary imagery. While \method as a map-free approach would not require accurate global reference poses as provided by Mapillary, this was the only publicly available service to expose the required information, \ie~posed reference images with intrinsic parameters.

\begin{figure}[h]
        \centering
        \begin{tabular}{cc}
            \textbf{Initial localization} & \textbf{Localization 1} \\
             \includegraphics[width=.5\columnwidth]{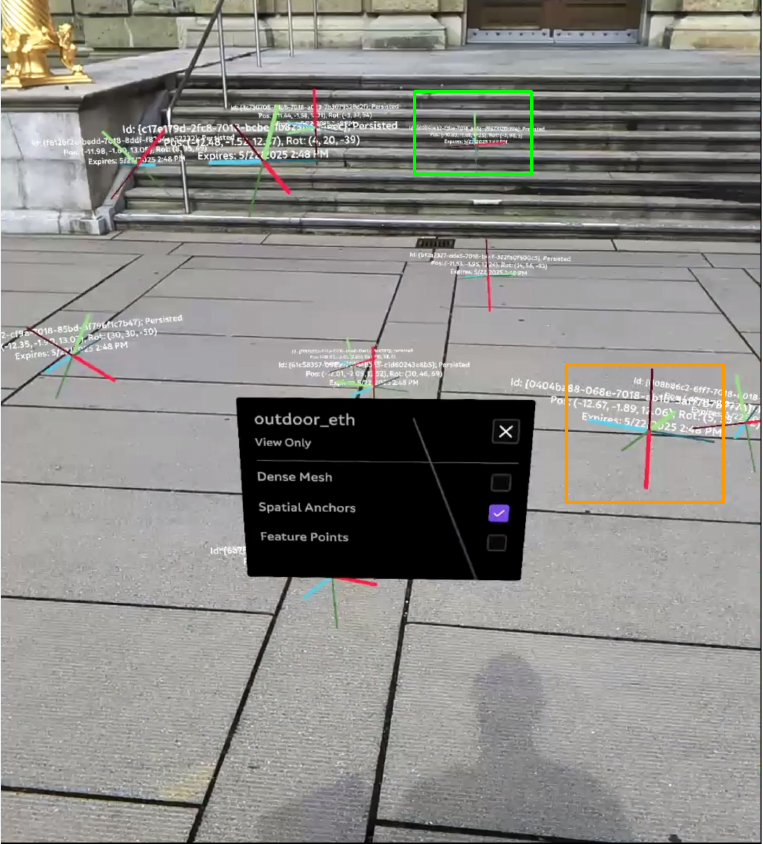} & \includegraphics[width=.5\columnwidth]{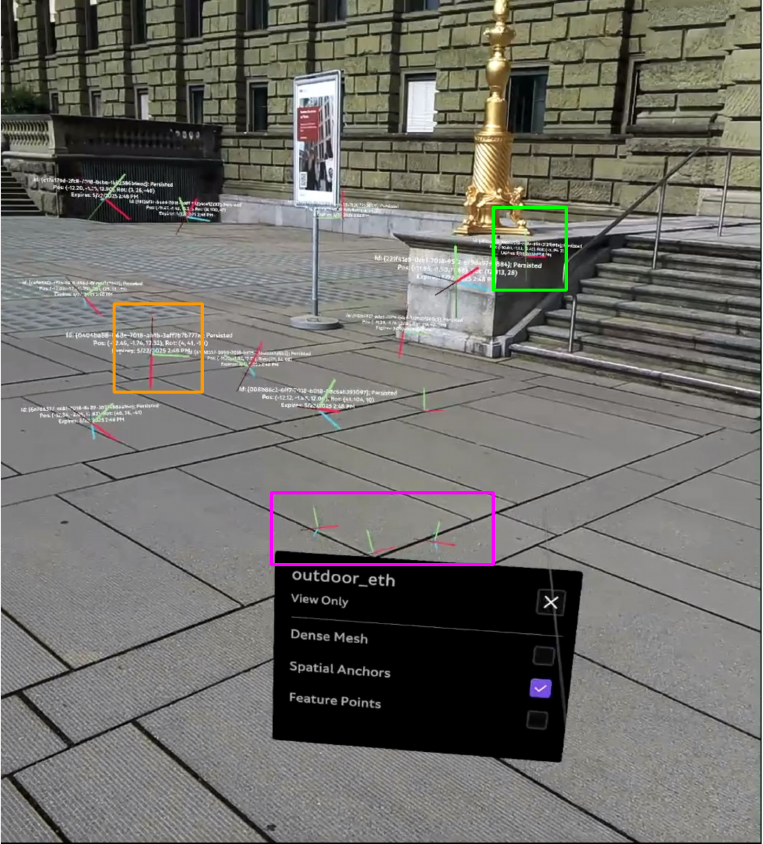} \\
             \textbf{Localization 2} & \textbf{Localization 3} \\
             \includegraphics[width=.5\columnwidth]{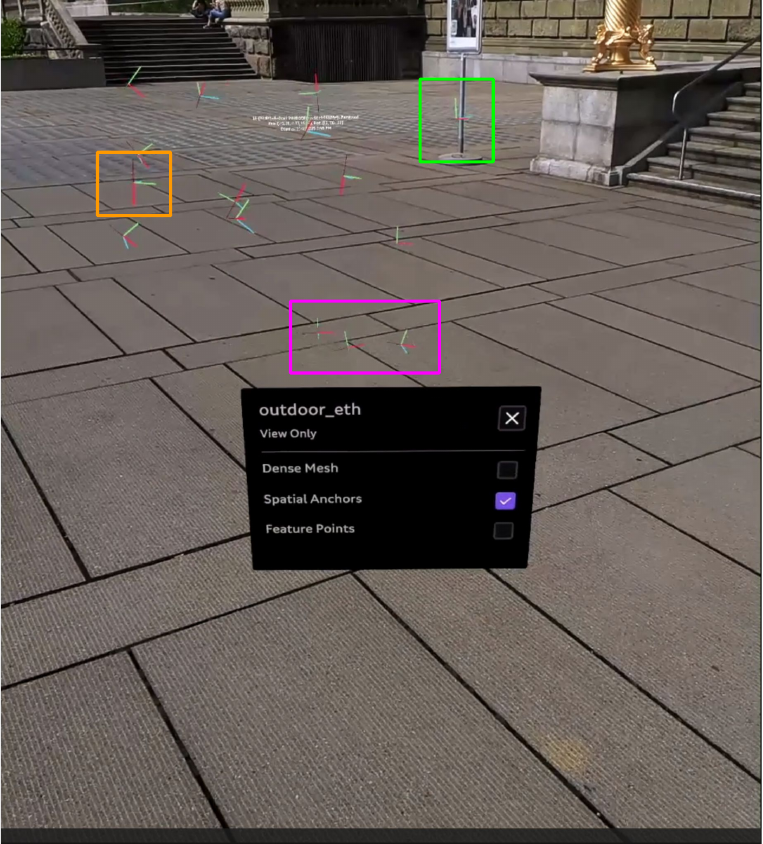} &\includegraphics[width=.5\columnwidth]{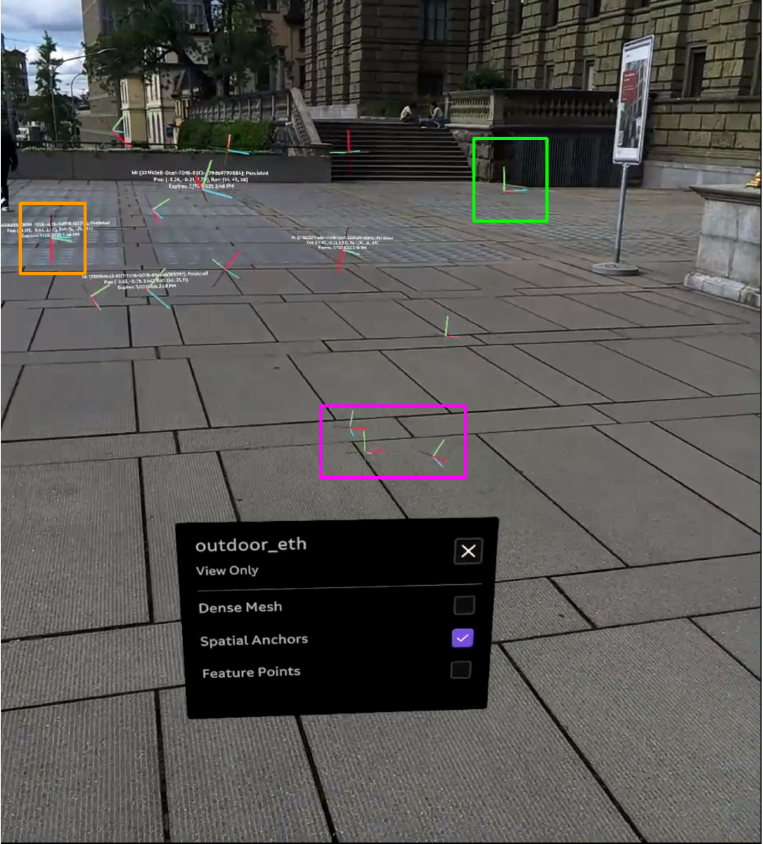}
        \end{tabular}
        \caption{Visual content after several localization events using our method integrated into an ML2 device. The visual content is placed after the initial localization. After each localization event, the visual content remains close to its original position in the physical space. The bounding boxes distinguish different visual content at each localization event.}
    \label{fig:ml_test}
\end{figure}

\mypar{Qualitative evaluation} In the absence of accurate ground-truth data in this setup, we evaluate the performance of our method with the ML2 device using the following procedure: (i) First, we employ our method to localize the ML2 within the global frame of reference. (ii) Next, we place example visual content (in our experiments a virtual 3D coordinate frame) at various distinguishable physical locations within this global frame based on the initial localization event. (iii) Finally, we use our method to relocalize the ML2 with respect to the global frame from a different point of view. Accurate localization is indicated if the visual content remains close to its initial position. \figref{fig:ml_test} showcases a qualitative evaluation of our method using the described setup. As presented, the visual content (3D coordinate system) remains in a similar position throughout the three localization events. The colored bounding boxes highlight the same selected virtual content after successive localization events throughout the experiment. In addition to \figref{fig:ml_test}, the video~\footnote{\url{https://youtu.be/972XoLbXkYQ}} accompanying this article shows the experiment described in this section.

\setlength{\tabcolsep}{1pt}

%% file: 6-conclusions.tex
\section{Conclusions}\label{sec:conclusions}
In this letter we introduce a novel map-free visual localization framework for AR applications - namely \method. Our approach leverages the known relative transformations of the query sequence to perform intra-sequence triangulation and lift 2D-2D matches into the 3D space, followed by pose estimation with \gls{pnp} and further refinement with \gls{pgo}. Our evaluation on benchmark datasets for map-free and map-based localization showcases the competitive performance of \method compared to other \sota methods addressing the map-free case. Furthermore, \method achieves results within the same range as traditional SfM-based localization approaches, which are the gold standard in visual localization, but without requiring the use of a time- and compute-intensive mapping phase. Finally, we test our method in a real-world experiment using a Magic Leap 2 AR device and publicly available map data from Mapillary, attesting to the practicality of \method for \gls{ar} use-cases.